\documentclass[letterpaper, 10pt, journal, twoside]{IEEEtran}  
\IEEEoverridecommandlockouts
\usepackage{cite}
\usepackage{amsmath,amssymb,amsfonts}
\usepackage{graphicx}
\usepackage{textcomp}
\usepackage{xcolor}
\usepackage{basicmath}
\usepackage[caption=false, font=footnotesize]{subfig}
\usepackage[normalem]{ulem}
\usepackage{hyperref}

\DeclareMathOperator*{\argmin}{arg\,min}

\def\BibTeX{{\rm B\kern-.05em{\sc i\kern-.025em b}\kern-.08em
    T\kern-.1667em\lower.7ex\hbox{E}\kern-.125emX}}
\begin{document}


\newcommand{\Will}[1]{\textcolor{blue}{#1}}
\newcommand{\Brian}[1]{\textcolor{red}{#1}}


\title{ Validating Robotics Simulators on Real-World Impacts}

\author{Brian Acosta*, William Yang*, and Michael Posa
\thanks{This material is based upon work supported by the National Science Foundation Graduate Research Fellowship Program under Grant No. DGE-1845298. Toyota Research Institute also provided funds to support this work.}
\thanks{*These authors contributed equally. Names are in alphabetical order.}
\thanks{The authors are with the GRASP Laboratory, University of Pennsylvania, Philadelphia, PA 19104, USA {\tt\footnotesize\{bjacosta, yangwill, posa\}@seas.upenn.edu}}
\thanks{© 2022 IEEE.  Personal use of this material is permitted.  Permission from IEEE must be obtained for all other uses, in any current or future media, including reprinting/republishing this material for advertising or promotional purposes, creating new collective works, for resale or redistribution to servers or lists, or reuse of any copyrighted component of this work in other works}
}

\maketitle

\begin{abstract}
A realistic simulation environment is an essential tool in every roboticist's toolkit, with uses ranging from planning and control to training policies with reinforcement learning. 
Despite the centrality of simulation in modern robotics, little work has been done comparing robotics simulators against real-world data, especially for scenarios involving dynamic motions with high speed impact events. 
Handling dynamic contact is the computational bottleneck for most simulations, and thus the modeling and algorithmic choices surrounding impacts and friction form the largest distinctions between popular tools.  Here, we evaluate the ability of several simulators to reproduce real-world trajectories involving impacts. 
Using experimental data, we identify system-specific contact parameters of popular simulators Drake, MuJoCo, and Bullet, analyzing the effects of modeling choices around these parameters. 
For the simple example of a cube tossed onto a table, simulators capture inelastic impacts surprisingly well, though generally fail to reproduce elasticity. For the higher-dimensional case of a Cassie biped landing from a jump, the simulators capture the bulk motion well but the accuracy is limited by model differences between the real robot and the simulators.
    
\end{abstract}

\begin{IEEEkeywords}
Contact Modeling, Simulation and Animation, Dynamics
\end{IEEEkeywords}

\section{Introduction}
Given the importance of simulation in planning and control, it is important to understand the physical realism of simulated impacts. Recent successes in sim-to-real reinforcement learning for legged locomotion \cite{siekmann_sim--real_2021}, \cite{kumar_rma_2021}, \cite{xie2020learning} largely use domain randomization to mitigate model uncertainties. However, domain randomization works best when the range of randomized model parameters is small, and non-randomized dynamics are accurate. Contact parameters other than friction are rarely randomized, suggesting  accurate impact simulation could be a contributing factor in successful sim-to-real transfer. 
Impact simulation accuracy also affects the ability to accurately compute regions of attractions for legged robot control \cite{ubellacker_verifying_2021} and verify the robustness of impact-aware control techniques \cite{yang_impact_2021}, \cite{gong2020angular} without risking failure.
Understanding which impact behaviors can be recreated in popular robotics simulators is an important step in applying these techniques to more impact-rich behaviors.

Most common robotics simulators use approximately rigid contact models \cite{stewart_and_trinkle},\cite{anitescu1997formulating}, with specific approximations chosen for the sake of computational speed and numerical stability. 
The physical realism of these contact models are largely validated by visual inspection or by tangential physical metrics \cite{erez_simulation_2015}, and there is evidence that rigid contact models sometimes poorly predict the dynamics of real collisions even for single impacts \cite{fazeli_empirical_2017}, \cite{chatterjee_realism_1999}. 
Follow up work on improving the prediction ability of rigid body models with residual learning \cite{fazeli_long-horizon_2020} demonstrates some success, though recent work suggests that stiff contact behavior may lead to poor amenability to learning based techniques \cite{parmar_fundamental_2021}.


\begin{figure}[t]
	\centering
	\includegraphics[width=0.45\textwidth]{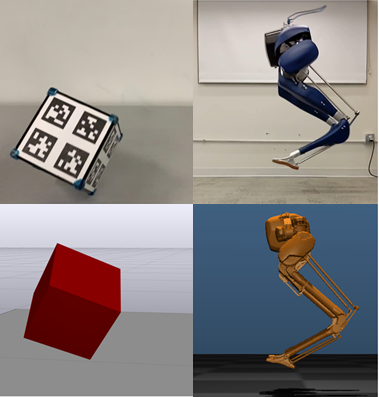}
	\caption{Athletic behaviors such as jumping (right) involve high speed impacts which are difficult to model and simulate. Yet, optimal performance of controllers which are trained or verified in simulation relies on such impacts being faithfully captured by simulators. In this work, we evaluate simulators' ability to capture impact dynamics using real-world data from Cassie jumping and a cube tossed onto a wooden table.}
	\label{fig:sim_initial_conditions}	
\end{figure}

While simulators may be able to accurately capture real impacts if tuned to a correct set of parameters, roboticists often choose contact parameters such as stiffness to maximize simulation speed and may only be peripherally aware of the tradeoffs being made with physical realism. 
As part of this paper, we examine the consequences of this approach by conducting a sensitivity analysis of simulators' physical realism to the contact parameters.

The primary contributions of this work are:
\begin{itemize}
\item Empirical evaluation of multiple simulators on complex 3D impacts using real world data collected from cube tosses and jumping with a bipedal robot.
\item Identification of optimal contact parameters for each system in Drake, MuJoCo, and Bullet and sensitivity analysis to assess the importance of properly specifying each parameter.
\item Analysis of the relative strengths and weaknesses of each simulator when applied to the given scenarios. 
\item A publicly available dataset containing Cassie jumping trajectories. 
\end{itemize}

\section{Background}
\label{section:background}

The forward simulation of rigid body motion without contact is assumed to be identical across all three simulators. For this reason, this paper focuses on the predictive fidelity of simulator contact models when the system undergoes impact. Contact models define a mathematical relationship between the relative position and velocity of two rigid bodies and the contact force between them.  Rigid contact models assume that bodies are perfectly rigid (i.e. infinite stiffness) and undergo inelastic collisions, which can be posed as a linear complementarity problem (LCP) \cite{stewart_and_trinkle}. Compliant models acknowledge that real objects experience some deformation during contact, and approximate the inter-body forces produced by deformation as a restoring force against the interpenetration of objects. The constitutive laws of compliant contact models are numerically stiff, requiring the use of small time steps, while solving LCPs is computationally expensive. Therefore, simulators take different approaches to improving simulation speed by choosing approximations and/or solution tactics. The following section details the contact model and solution strategy for each of the chosen simulators, which represent three distinct strategies for modeling and solving for contact forces.


\subsection{Simulator Contact Models}
\subsubsection{Drake}
Drake \cite{drake} uses a compliant model of contact with Hunt \& Crossley dissipation \cite{hunt_coefficient_1975} which penalizes interpenetration using a nonlinear spring-damper law. Using a smooth approximation of Coulomb friction, Drake expresses contact forces as a function of state. Given a normal penetration distance $\delta$ and penetration rate $\dot{\delta}$, the normal contact force is 
 \begin{align}
     f_{n} = k(1 + b \dot{\delta})_{+} \delta_{+},
 \end{align}
 where $k$ is the stiffness, $b$ is the dissipation, and $(\cdot)_{+} = \max(0, \cdot)$ ensures positive normal forces. When incorporated into the equations of motion, this leads to a 
 nonlinear system of equations. Drake uses the custom TAMSI solver to properly
resolve contact transitions \cite{castro2020transition}. Like many other contact-model solvers, TAMSI uses an iterative algorithm; however, TAMSI enforces convergence at each time step. Resulting contact forces are therefore consistent with the original modeling equations of compliant contact with regularized friction. 
    
\subsubsection{MuJoCo}
\label{subsec:mujocobackground}
MuJoCo uses a convex approximation of rigid contact first introduced in
\cite{anitescu_optimization-based_2006} with regularization introduced in \cite{todorov_convex_2014}. 
This regularization softens contact by relaxing the strict complementary of contact force and distance between objects. The amount of regularization is determined by user specified stiffness and damping parameters, which are transformed internally into reference accelerations for the contact constraint dynamics, which approximately obey the following linear spring-damper law \footnote{\url{https://mujoco.readthedocs.io/en/latest/modeling.html}}:
\begin{align}
a_{n} \approx (-k\delta - b\dot{\delta})d(\delta) + (1 - d(\delta)) a_{0} , 
\label{eq:mujoco_contact_model}
\end{align}
where $a_{n}$ is the normal acceleration, $a_{0}$ is the acceleration in the absence of contact, and $d(\delta) \approx 1$ is a position-dependent interpolation between the constrained and unconstrained acceleration. MuJoCo allows for tuning the shape of $d$, though the results presented here use the default values, as we find no improvement in the prediction error by tuning $d$.
MuJoCo implements friction by solving a convex optimization problem which balances achieving the constraint dynamics in (\ref{eq:mujoco_contact_model}) with contact activation and energy dissipation. 

This convex formulation guarantees a unique solution at each time step and yields fast, differentiable computations. 
This speed has made MuJoCo a popular simulator in the robotics community, especially among RL practitioners, though MuJoCo's contact constraint regularization has been observed to cause non-physical artifacts during slip. Objects can glide at a distance from each other \cite{mazhar_analysis_2015}, and large amounts of regularization can lead to viscous slip \cite{noauthor_simbenchmark_nodate}, requiring the use of additional stabilization techniques. Like Drake, after posing its regularized dynamics problem, MuJoCo is guaranteed to find an accurate solution to this problem. 

\subsubsection{Bullet}
Bullet is a physics engine originally introduced for simulation in computer graphics and animation but widely used for robotics simulations. Bullet formulates the contact problem as a mixed LCP (MLCP), a generalization of the LCP which allows for inequality constraints. Bullet's MLCP represents rigid contact with Coulomb friction, and is solved using a Projected Gauss-Seidel (PGS) solver. PGS attempts to iteratively resolve each contact constraint separately, keeping the remaining constraints fixed. Unlike Drake or MuJoCo, Bullet's PGS solver is allowed to return with an intermediate computation when a fixed maximum number of
iterations is reached. In this case, Bullet relies on linear spring-damper Baumgarte stabilization \cite{BAUMGARTE19721} to enforce the contact constraint.
This feature can cause a lack of robustness in the event the PGS solver fails to converge.

\subsection{Simulator Considerations}
\label{subsec:simulator_reasons}
	
	The simulators evaluated in this paper are by no means exhaustive, with several other engines finding popularity in the robotics community.
	The three chosen here, however, provide good coverage of the different styles of contact dynamics formulations.
	We note that we considered IsaacGym \cite{makoviychuk2021isaac}, a popular simulator for reinforcement learning applications, including sim-to-real transfer on a legged robot \cite{rudin2022learning}.
	However, at the time of publication, IsaacGym does not support modification of its compliance parameters.
	Without this capability, the accuracy of the IsaacGym simulator was quite poor on our dataset.

	For the Cassie dataset, we limit our comparisons to Drake and MuJoCo.
	Bullet does not natively support modeling the reflected inertia from the motors and gearboxes, which plays a significant role in the dynamics of Cassie.
	
%

\section{Experimental Setup}

\subsection{Cube Toss}
\subsubsection{Data}
There are 550 trajectories of a 10 cm acrylic cube tossed onto a wooden table as a dataset for parameter identification and performance evaluation. The data collection procedure is described in  \cite{pfrommer2020contactnets} and the data is available as part of an open source code repository\footnote{\url{https://github.com/DAIRLab/contact-nets}}.  We name these the ground truth cube trajectories and denote a ground truth trajectory of length $T$ as $\vect{x}^{*} = (x_{t}^{*})_{t = 1 \ldots T}$ where $x_{t}^{*} = [q_{t}^{*}; v_{t}^{*}]$ is the state of the cube at each time step. The configuration $q$ consist of position $p \in \mathbb{R}^{3}$ and orientation $R \in SO(3)$, and the velocity $v$ consists of linear and angular velocities $\dot{p}$ and $\omega$. Physical properties of the cube-table system are given in Table \ref{tab:cube_physics}. The cube was weighed to determine the mass, with the inertia determined from the measured mass and geometry. Friction was determined via a tilt-test, and restitution was determined by minimizing the prediction error in \cite{pfrommer2020contactnets}. 

\begin{table}[h]
	\caption{Cube Physical Parameters}
	\begin{center}	
		\begin{tabular}{|c|c|c|c|}
			\hline
			\textbf{Mass} & \textbf{Inertia} & \textbf{Friction} & \textbf{Restitution}\\
			\hline
			0.37 kg & .0081 kg m\textsuperscript{2} & 0.18 & 0.125 \\
			\hline
		\end{tabular}
		\label{tab:cube_physics}
	\end{center}
\end{table}

\subsubsection{Simulation Environment}
Each cube toss simulation is designed to match the real experiment as closely as possible. The measured dimensions and inertial parameters of the real cube are specified in a Universal Robot Description Format (URDF) file for Bullet and Drake, and as XML text in the MJCF format for MuJoCo. The timestep for each simulator is set to 1480 Hz, and the resulting trajectories are down-sampled to match the data collection frequency of 148 Hz. We found that decreasing the simulator timestep further did not improve the prediction capability of any simulator.

\subsection{Cassie Jumping}
\subsubsection{Data Collection}
We use  state and input data from 22 jumping experiments performed with the Cassie bipedal robot.
The jumping trajectories were generated using the jumping controller detailed in \cite{yang_impact_2021}, and are available as a public dataset\footnote{\url{https://github.com/DAIRLab/cassie_impact_dataset}}.
The state measurements, sampled at 2000Hz, are composed of the joint positions and velocities as well as the floating base state of the pelvis.
Although we treat the state data as ground truth, there is uncertainty in these measurements as a state estimator \cite{hartley2020contact} is used to compute the floating-base state and joint velocities are subject to encoder noise and resolution.
To account for state estimation errors, we offset the pelvis vertical height so that the feet are in contact with the ground at impact. 
This ensures that the impact timing between the real and simulated data matches. 
For context, the maximum correction applied is $0.02m$, which corresponds to approximately 20\% of the total loss for both Drake and MuJoCo.

We use the same notation for Cassie trajectories as with the cube, though Cassie is modeled as a floating-base Lagrangian system with $q \in \mathbb{R}^{n+7}$ and $v \in \mathbb{R}^{n + 6}$ where $n = 16$ is the number of joints.
The motor input data, also sampled at 2000Hz, are the torques measured at the motors as opposed to the torques commanded by the controller.
We make this distinction to account for the delays between the output of the controller and when the motor actually is supplied the commanded current.

\subsubsection{Simulation Environment}

Simulated data is generated from Drake and MuJoCo, and Bullet is not used on the Cassie data for reasons explained in \ref{subsec:simulator_reasons}.

The description of the robot is specified in a URDF for Drake and as a XML for MuJoCo.
For both simulators, we initialize the state of the robot just prior to impact and apply the measured motor torques at the corresponding times.
This decision to simulate in open-loop eliminates the dependency on the controller.
For both simulators, the state was sampled at 2000Hz, consistent with the hardware data.

\begin{figure}[h]
    \centering
    \includegraphics[width=0.42\textwidth]{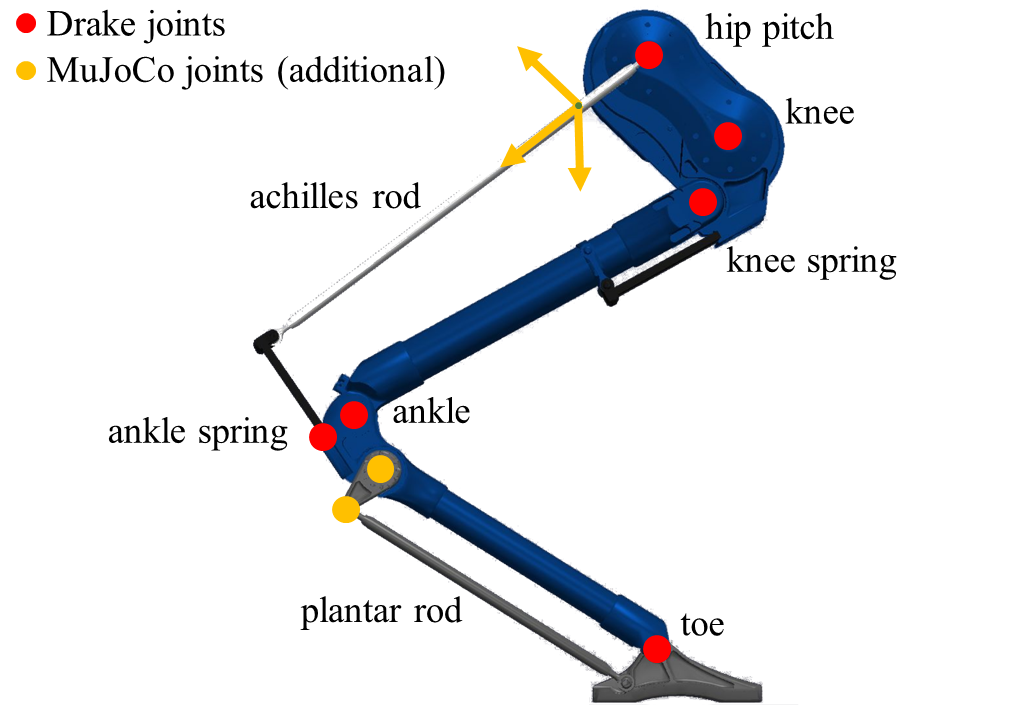}
    \caption{The configuration states for a leg of the Cassie robot. The additional joints included in the MuJoCo model are indicated in yellow, which include the orientation of the achilles rod and the angles of the foot crank linkage.}
    \label{fig:cassie_kinematics}
\end{figure}

There are slight differences in the Cassie models used by Drake and MuJoCo.
The physical Cassie robot contains two four-bar linkages per leg that enable the control of leg length through the knee motor and the control of the toe joint through an actuator located at the ankle.
The MuJoCo simulator provided by Agility Robotics \cite{agility_robotics_cassie-mujoco-sim_2018} includes the achilles and plantar rods shown in Fig. \ref{fig:cassie_kinematics} that form the loop closures for the linkages.

Drake does not natively support loop closure constraints, thus the URDF does not include the connecting rods but accounts for their inertial contribution by adding lumped masses at the anchor points.
The upper loop closure is modeled as a stiff spring with spring constant chosen to enforce the loop closure but not interfere with the performance of the simulator.
The lower loop closure is handled by applying actuator efforts directly at the toe joint, a common convention used in many Cassie models \cite{reher_dynamic_2019}, \cite{gong2019feedback}.

While the state of the connecting rods are fully defined by the other states, the MuJoCo model includes these redundant states.
For fair comparison, we map the full MuJoCo states to the Drake states when comparing state trajectories.

\section{Parameter Identification}
To evaluate the performance of each simulator, we identify a simulator and system-specific set of contact parameters that best enables each simulator to reproduce real-world trajectories.
To maintain some comparability across simulators, we only tune the friction, stiffness, and damping parameters for each simulator. The parameters and their physical significance are given in Table \ref{tab:param_candidates}. Beyond differences in units, these parameters may not be physically equivalent, as the contact laws vary between simulators. We combine static and dynamic friction into a single constant $\mu$ as we do not see higher accuracy from distinguishing between coefficients.

 \begin{table}[h]
    \centering
    \caption{Contact Parameters Identified}
    \begin{tabular}{|c|c|c|c|}
    \hline
        & \textbf{Parameter} & \textbf{units} & \textbf {Physical Interpretation} \\
    \hline
     & $\mu$ & - & Friction coefficient\\
     \textbf{Drake} & $k$ & N / m & Contact stiffness \\
     & $b$ & s / m & Contact dissipation\\
     \hline
     & $\mu$ & - & Friction coefficient\\
     \textbf{MuJoCo} & $k$ & N / (kg m) & Contact stiffness \\
     & $b$ & N s / (kg m) & Contact damping \\ 
     \hline
      & $\mu$ & - & Friction coefficient\\ 
     \textbf{Bullet}& $k$ & N / m & Contact stiffness\\
     & $b$ & N s / m & Contact damping \\
     \hline
    \end{tabular}
    \label{tab:param_candidates}
\end{table}

\subsection{Evaluation Metrics}
 For a ground truth trajectory $\vect{x}^{*}$, the corresponding simulator trajectory $\vect{\hat{x}}(x_{0}^{*}, \theta) = (\hat{x}_{t})_{t = 1 \ldots T}$, is simulated with contact parameter vector $\theta$ from the initial condition $x_{0}^{*}$. We define the following error metrics, which will be used in the loss function for parameter identification. From here on we omit the dependence of $\vect{\hat{x}}$ on $x_{0}^{*}$ and $\theta$ for brevity.

\subsubsection{Cube Metrics}
The cube configuration error is 
\begin{multline}
    e_{q}(\vect{x}^{*}, \vect{\hat{x}}) = \\ \frac{1}{T} \sum_{t = 1}^{T} \big(\frac{2}{l}\TwoNorm{p^{*}_{t} - \hat{p}_{t}}^{2} +  \text{Angle}(R_{t}^{*}, \hat{R}_{t})^2 \big),
    \label{eq:e_q}
\end{multline}
where $l$ is the side length of the cube and $\text{Angle}(R_{1}, R_{2})$ is the angle of rotation of the relative rotation between $R_{1}$ and $R_{2}$. We scale the position error by $2 / l$ to give identical units and similar magnitudes to position and orientation error. Since the velocity in the cube toss data set is generated by filtering differences of positions, and is therefore influenced by filter dynamics, we do not calculate velocity errors, focusing instead on long term position and orientation accuracy. Due to the second order dynamics of the cube, this will capture the effect of finding the correct contact dynamics while not biasing toward incorrect velocity estimates in the ground truth data. 

\subsubsection{Cassie Metrics}
The joints on the Cassie robot have a wide range of inertias;
therefore, a naive loss function, such as the L2-norm, would lead to over-weighting of the low-inertia joints such as the toes.
To direct the parameter identification algorithm to prioritize capturing the bulk motion of the robot, we use a weighted norm
\begin{align}
    e_{cassie}(\vect{x}^{*}, \vect{\hat{x}}) = \sum_{t = 1}^T \tilde{x}_t^T W \tilde{x}_t,
\end{align}
where $W$ is a diagonal matrix, and $\tilde{x}$ is the error between the simulated and measured state.
The elements of $W$ are all $10$ for the position indices. 
For the velocity indices, we use a weight of $5$ for the floating base rotation, $100$ for the floating base translations, $0.01$ for the hip roll, knee spring, and toe joints, and $1$ for the remaining joints.
There are no measurements for the velocity of the ankle spring deflection, which is therefore omitted from the loss function.

\label{sec:cassie_model_differences}

    The data indicates significant but unknown model differences between the simulators and the physical robot.
    These differences may include incorrect model parameter values for the spring constants and joint damping.
    Better modeling these effects would likely improve simulation accuracy, but the complexity of Cassie makes full system identification infeasible and is outside the scope of this paper.
    
    To mitigate the effects of model uncertainty mentioned above, and to focus on capturing the impact event, we evaluate the trajectories for a brief 50 ms time window (T = 100 when sampled at 2000 Hz) around the impact event.
    This is under the assumption that the contact forces are large and the time window is short enough that error from incorrect model parameters will have a relatively small effect.
    To account for timing variations between jumping experiments, we manually select the time window for each log to include the first impact event for each jump.

\subsection{Optimization Procedure}
\begin{table}[h]
	\centering
	\caption{Contact Parameter Optimization Domains}
	\begin{tabular}{|c|c|c|c|}
		\hline
		& \textbf{Parameter} & \textbf{$\Theta_{cube}$ } & \textbf {$\Theta_{cassie}$} \\
		\hline
		\textbf{All Simulators} & $\mu$ & [0, 1] & [0, 1]\\
		\hline
		\textbf{Drake} & $k$ & [1e2, 1e5] & [1e3, 1e6]\\
		& $b$ & [0, 2] & [0, 3]\\
		\hline
		\textbf{MuJoCo} & $k$ & [1e2, 1e4] & [0,1e6] \\
		& $b$ & [0, 1e3] & [0, 1e3] \\ 
		\hline
		\textbf{Bullet}& $k$ & [1e2, 1e4] & -\\
		& $b$ & [0, 1e3] & - \\
		\hline
	\end{tabular}
	\label{tab:param_dom}
\end{table}

    We identify contact parameters using NGOpt, a gradient-free meta-optimizer provided through the \texttt{nevergrad} Python library \cite{nevergrad}. The appropriate contact parameters are the solution to the optimization problem
    \begin{align}
        \theta^{*} = \argmin_{\theta \in \Theta} L(\theta).
    \end{align}
    The optimization domains are shown in Table \ref{tab:param_dom}. The loss functions for the cube and Cassie are
    \begin{align}
        L_{cube} = \frac{1}{N} \sum_{i \in I} e_{q} (\vect{x}_{i}^{*}, \vect{\hat{x}}_{i})
    \end{align}
    and
    \begin{align}
        L_{cassie} = \frac{1}{N} \sum_{i \in I} e_{cassie}(\vect{x}_{i}^{*}, \vect{\hat{x}}_{i}),
    \end{align}
    where $I$ is the dataset and $N$ is the total number of trajectories in the dataset.

    
\section{Results}

\begin{figure*}[h]
	\centering
	\vspace{0.1in}
	\includegraphics[width=0.32\textwidth]{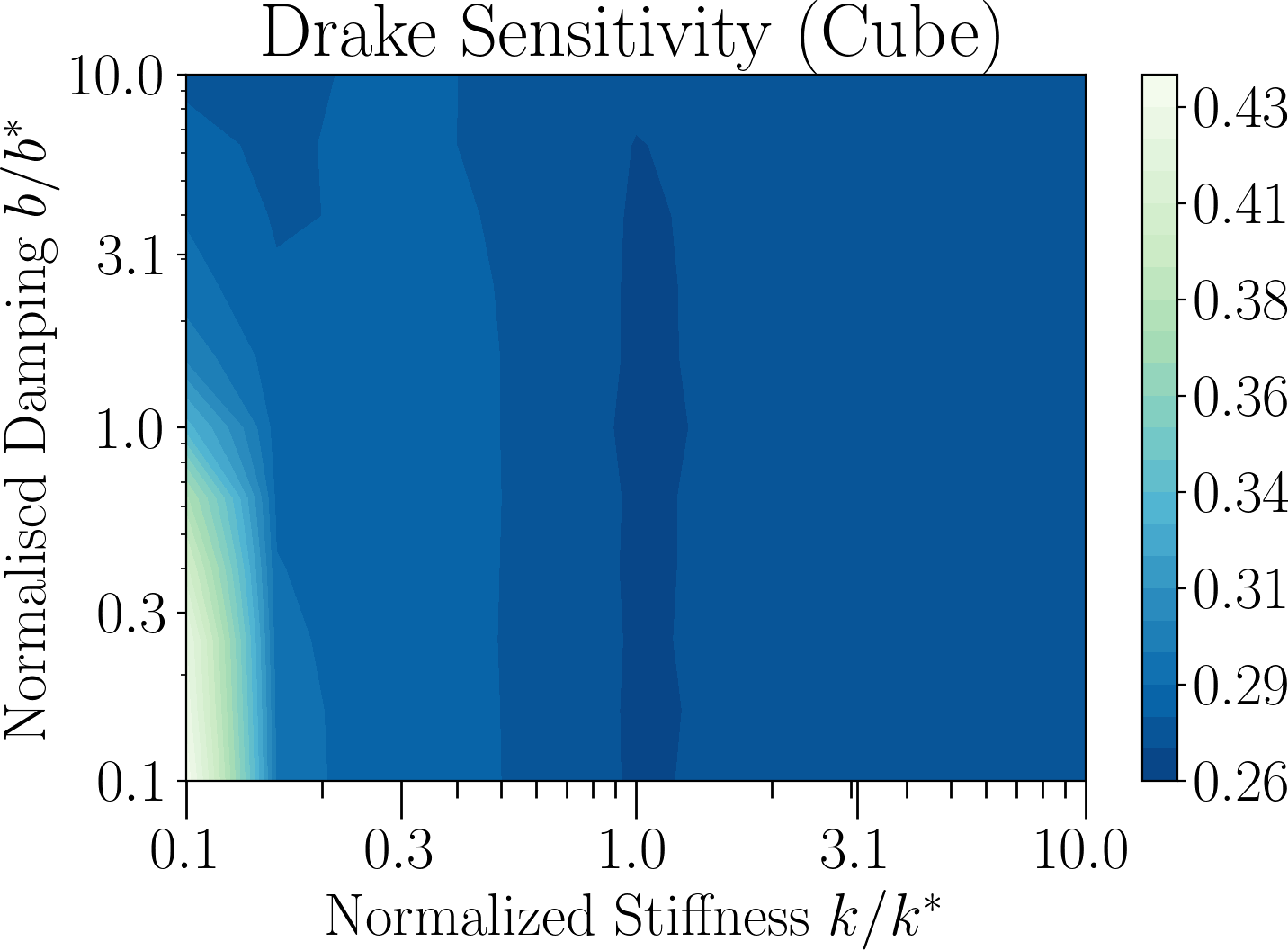}
	\includegraphics[width=0.315\textwidth]{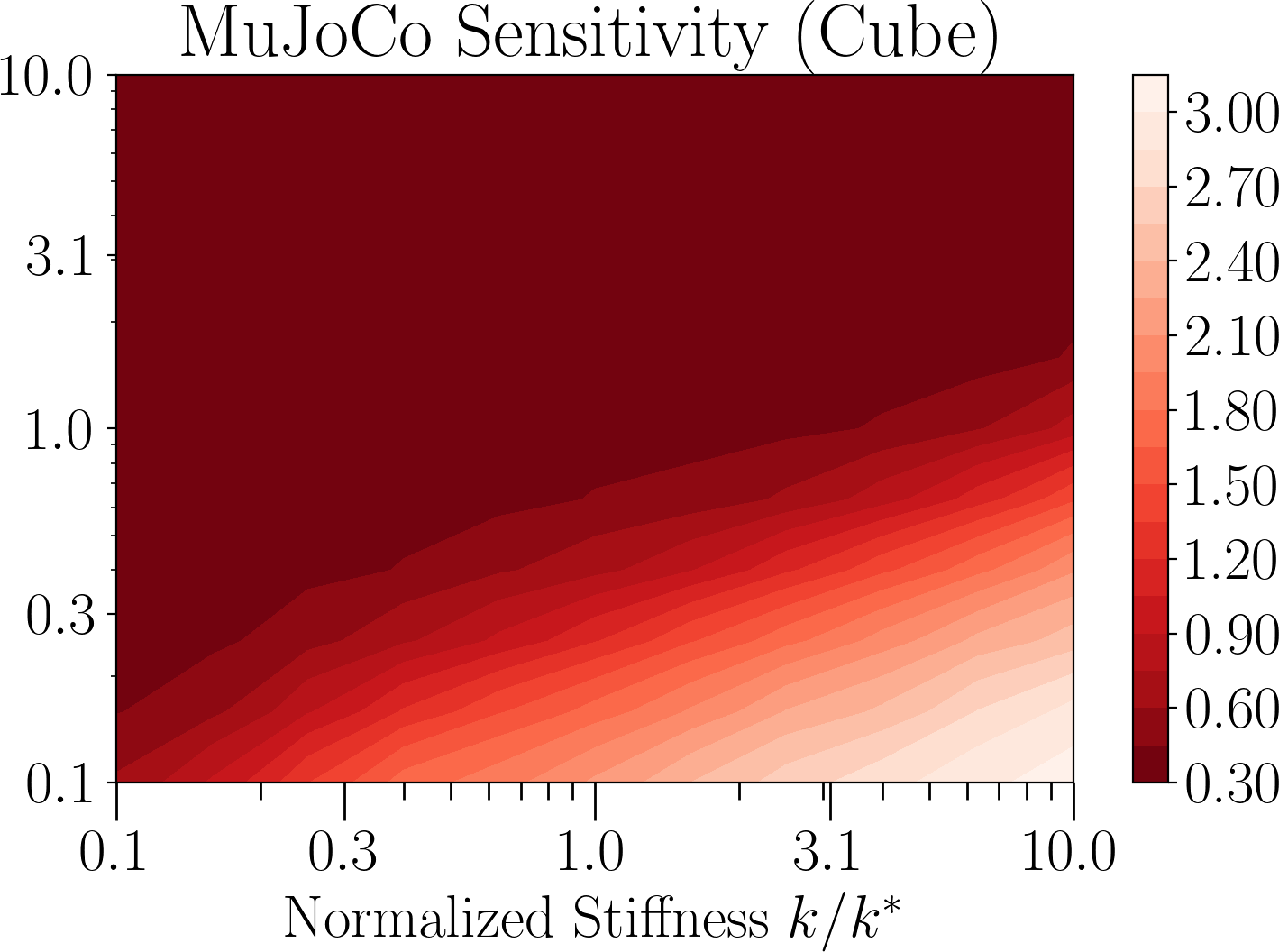}
	\includegraphics[width=0.32\textwidth]{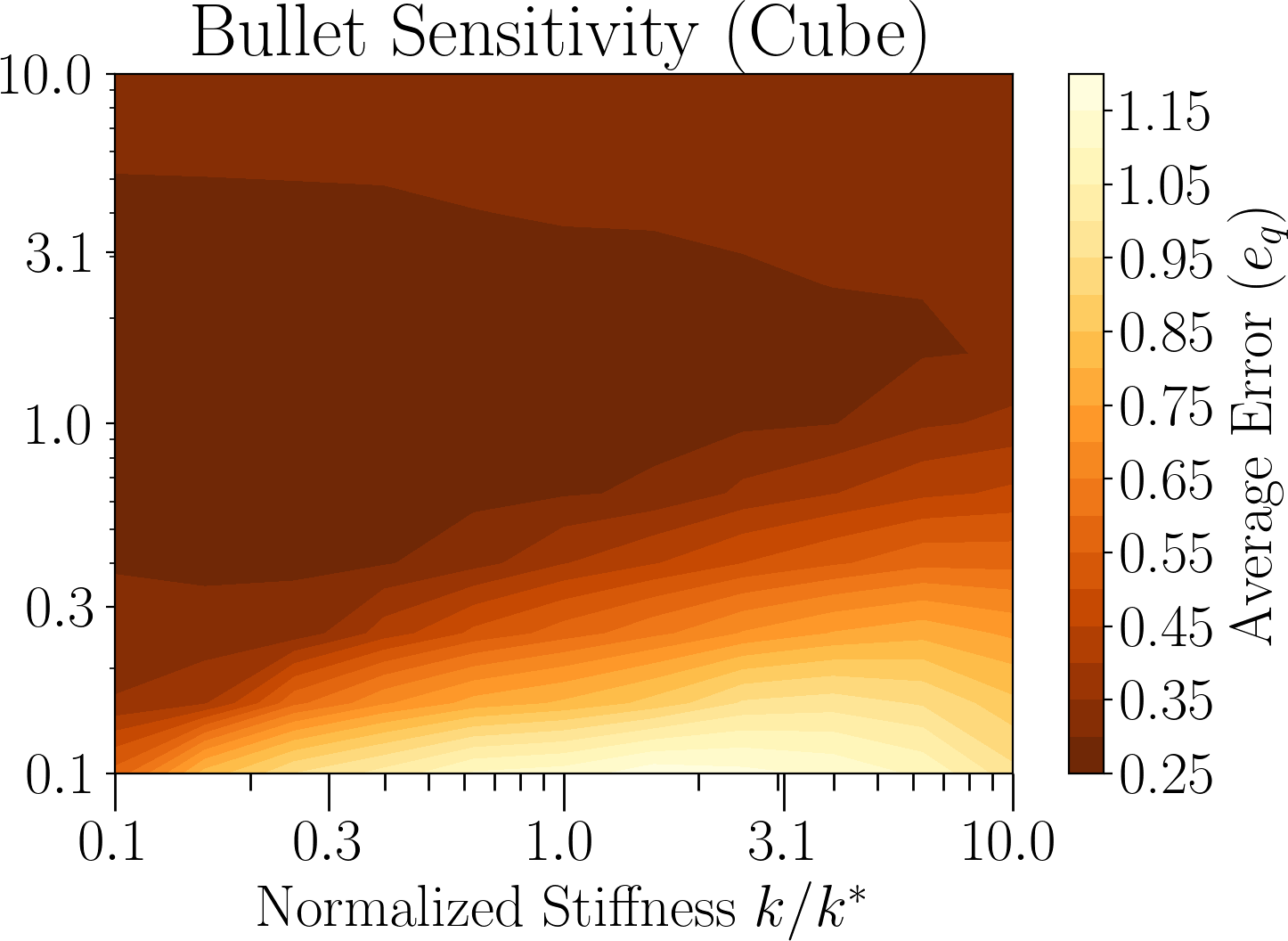}\\

	\includegraphics[width=0.32\textwidth]{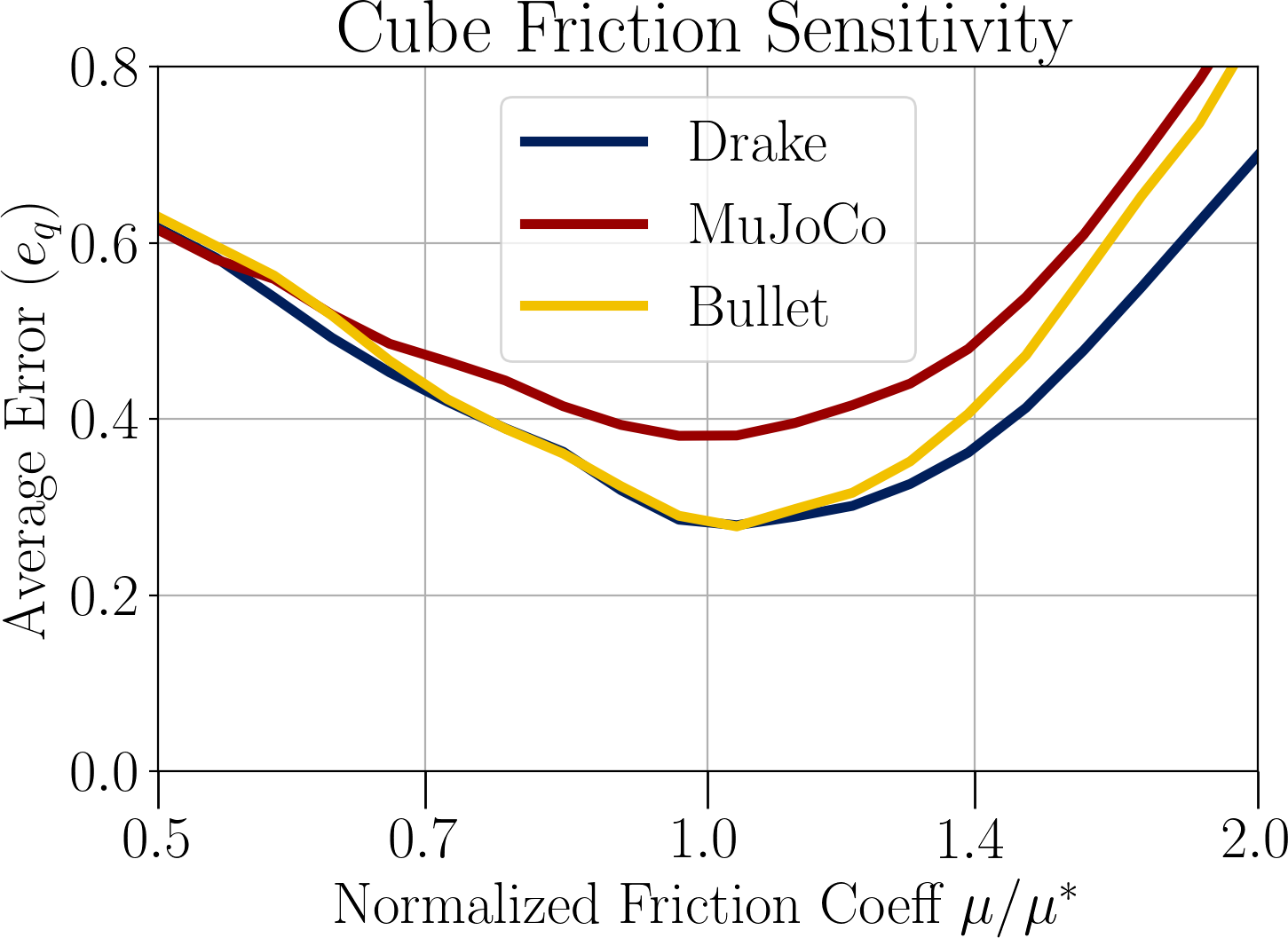}    
	\includegraphics[width=0.32\textwidth]{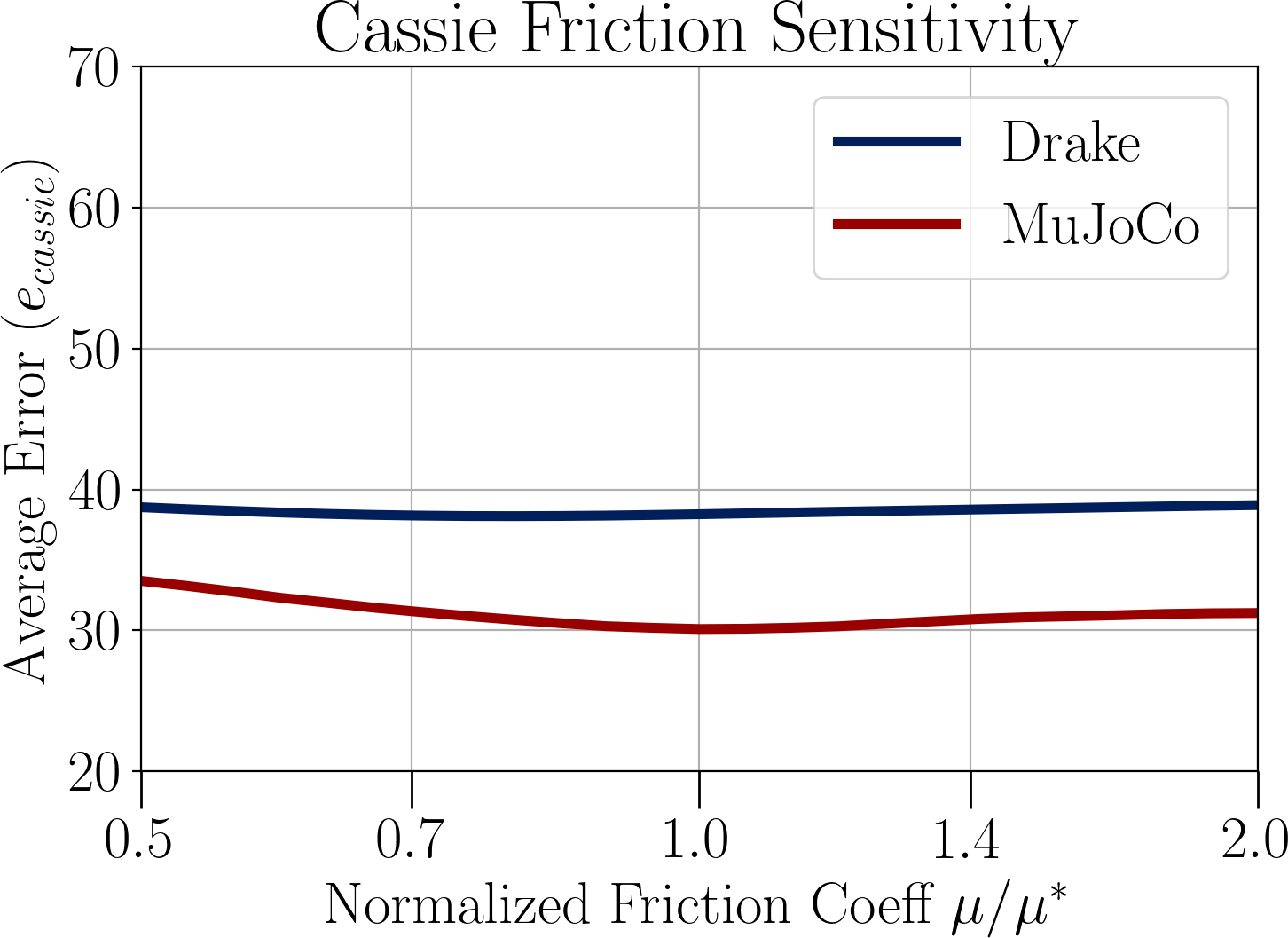}
	
	\vspace{.05in}
	\includegraphics[width=0.32\textwidth]{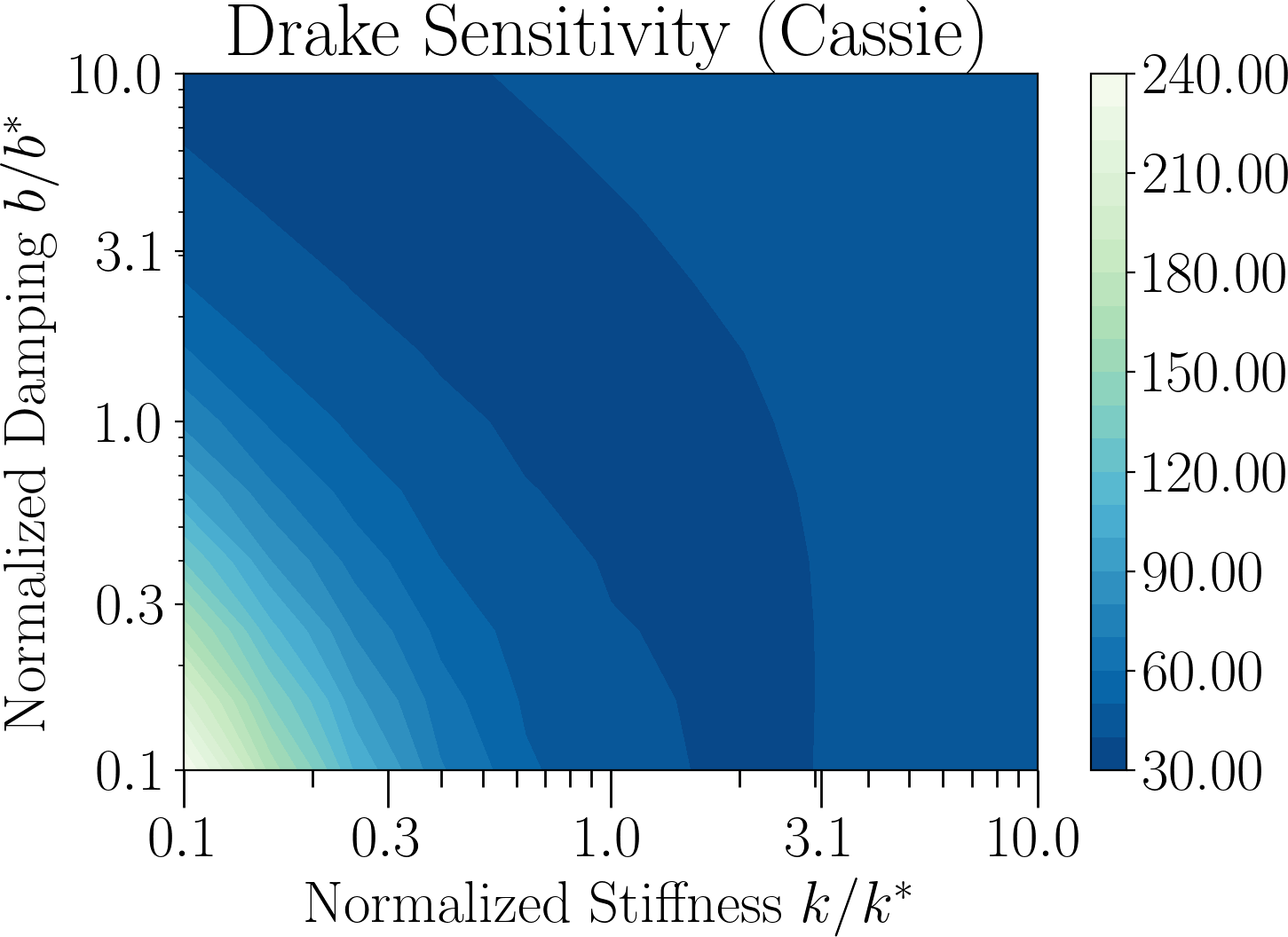}
	\includegraphics[width=0.32\textwidth]{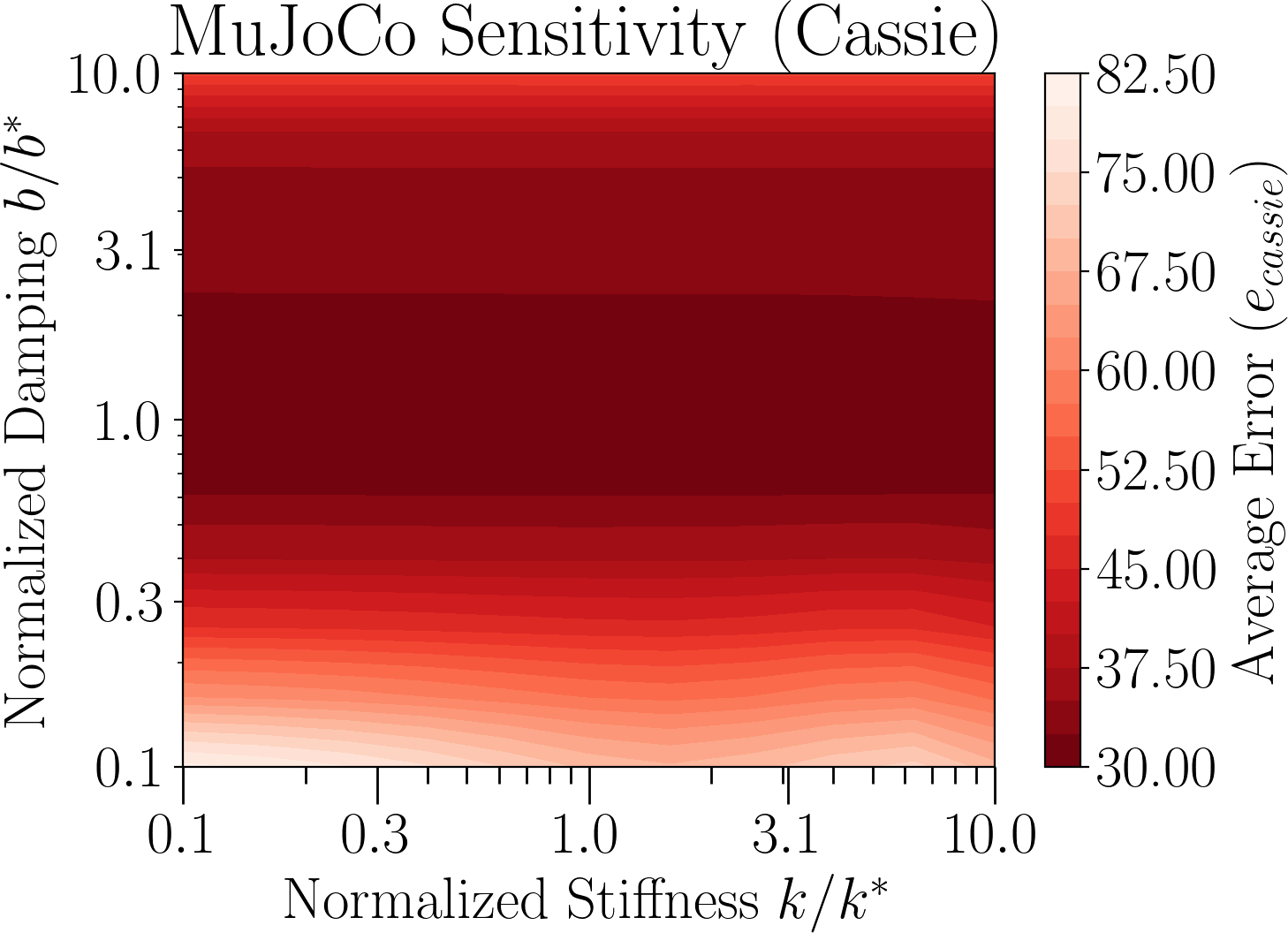}\\

	\caption{We perform a sensitivity analysis by holding friction fixed at the optimal value while sweeping a range of stiffness and damping values, and vice-versa. \textit{\textbf{Cube Toss} (Top Row):} All three simulators are mostly insensitive to stiffness above a given threshold, while MuJoCo and Bullet require sufficient damping as well. \textit{\textbf{Friction} (Middle):} Cube toss error is sensitive to the friction coefficient parameter due to large amounts of sliding in the dataset, while Cassie jumping is not, due to mostly experiencing static friction. \textit{\textbf{Cassie Jump} (Bottom):} Drake and MuJoCo display relatively low sensitivity to the contact parameters and have a wide range of parameter values that achieve low error. MuJoCo shows almost no sensitivity to its stiffness parameter. \vspace{-0.2in}}
	\label{fig:cube_sensitivity}
\end{figure*}

\subsection{Cube}
\subsubsection{Parameters}
The optimal contact parameters $\theta^{*}$ for each simulator are reported in Table \ref{tab:cube_params}\footnote{Due to a transcription error, incorrect Bullet parameters were reported in the original IEEE manuscript. The correct parameters are reported here. This error does not affect any other results or conclusions.}.
\begin{table}[htbp]
\caption{Identified Cube Toss Parameters}
\begin{center}
\begin{tabular}{|c|c|c|c|}
\hline
 &  &  \textbf{Dissipation/} & \\
 & \textbf{Stiffness} &\textbf{Damping} &  \textbf{Friction}\\ 
\hline
\textit{\textbf{Drake}} & 10800 & 0.4 & 0.10 \\
\hline 
\textit{\textbf{MuJoCo}} & 3300 & 45 & 0.22 \\
\hline
\textit{\textbf{Bullet}} & 1800 & 27 & 0.36 \\
\hline
\end{tabular}
\label{tab:cube_params}
\end{center}
\end{table}

\subsubsection{Performance}
As shown in Table \ref{tab:cube_errors}, the simulators are all able to accurately reproduce cube toss trajectories, with Drake and Bullet being more accurate than MuJoCo. In addition to the minimum $e_{q}$ observed for each simulator, we report average position and rotation error along with standard deviation $\sigma$ for each metric.

\begin{table}[h]
\caption{Summary of Cube Errors}
\begin{center}
\begin{tabular}{|c|c|c|c|}
\hline
 & \textbf{Position Err. $\pm \vect{\sigma}$.} & \textbf{Rotation Err.}  $\pm \vect{\sigma}$ & $\vect{e_{q}}$  $\pm \vect{\sigma}$ \\
 & \textbf{(\% Cube Width)} & (\textbf{Degrees}) &  \\
\hline
\textit{\textbf{Drake}} & 13.5 $\pm$ 8.2 & 16.5 $\pm$ 20.0 & 0.27 $\pm$ 0.57\\
\hline
\textit{\textbf{MuJoCo}} & 25.1 $\pm$ 10.8  & 21.7 $\pm$ 21.4 & 0.38 $\pm$ 0.63 \\
\hline
\textit{\textbf{Bullet}} & 14.9 $\pm$ 8.9 & 16.5 $\pm$ 20.2 & 0.27 $\pm$ 0.57 \\
\hline
\end{tabular}
\label{tab:cube_errors}
\end{center}
\end{table}

\vspace{-0.2in}
\subsection{Cassie}
The optimal contact parameters $\theta^*$ and the average loss for Drake and MuJoCo are reported in Table \ref{tab:cassie_params}.
Although we report a single set of optimal parameters, we observe that a wide range of parameters for both simulators achieve similar performance as shown in Fig. \ref{fig:cube_sensitivity}.
The parameter set that works well for MuJoCo is a damping value near the optimal value, and depends very little on the stiffness parameter.
Drake, on the other hand, has a band of optimal parameters that shows a clear relationship between stiffness and dissipation.

\begin{table}[htbp]
    \caption{Identified Cassie Jumping Parameters}
    \begin{center}
    \begin{tabular}{|c|c|c|c|c|}
    \hline
    &  &  & \textbf{Dissipation} & \\
     & \textbf{Loss} &\textbf{Stiffness} & \textbf{/Damping} & \textbf{Friction}\\
    \hline
    \textit{\textbf{Drake}} &  38.0 & 9400 & 3.0 & 0.45 \\
    \hline
    \textit{\textbf{MuJoCo}} & 30.0 & 1600 & 270 & 0.43\\
    \hline
    \end{tabular}
    \label{tab:cassie_params}
    \end{center}
\end{table}
\vspace{-0.2 in}
\section{Discussion}
Drake, MuJoCo, and Bullet are all capable of recreating the bulk motion of the trajectories seen in our datasets. While the following section discusses potential areas for improvement, the salient point should be that modern robotics simulators are able to capture dynamic impacts with friction for a range of contact parameters. 

\label{section:discussion}
\begin{figure}[]
    \centering
   	\vspace{0.1in}
    \includegraphics[width=0.47\textwidth]{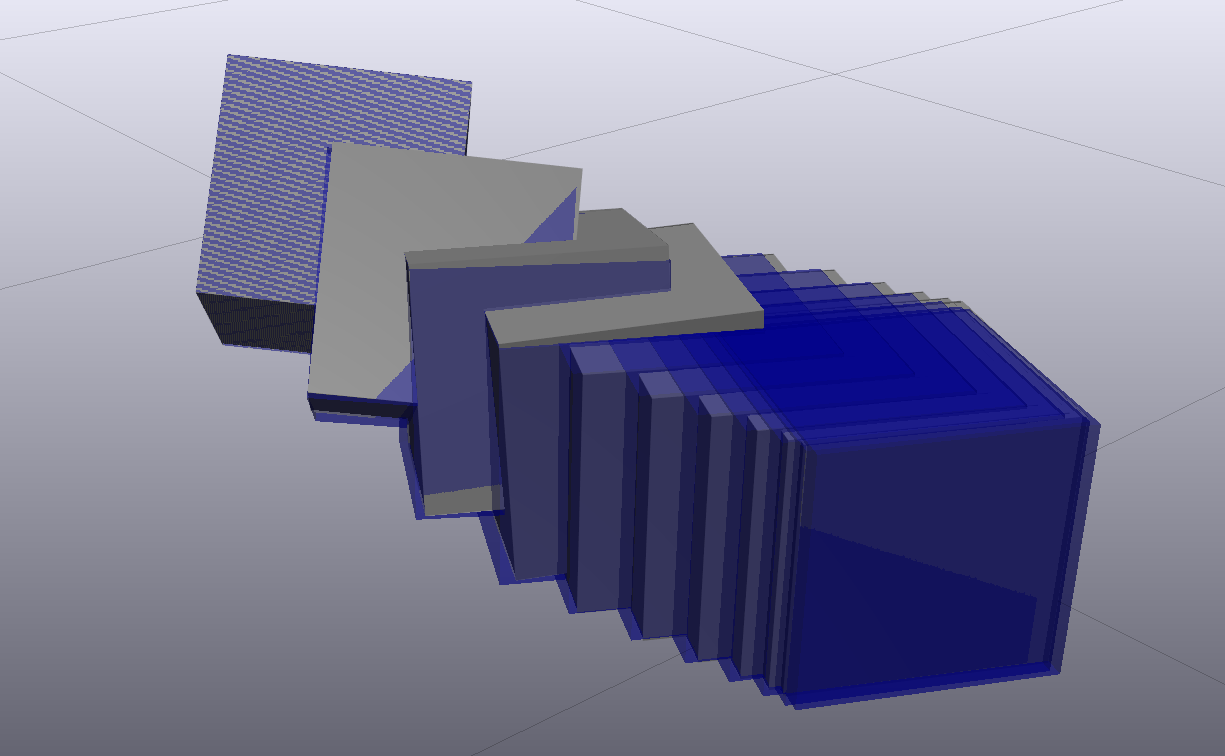}
    \includegraphics[width=0.47\textwidth]{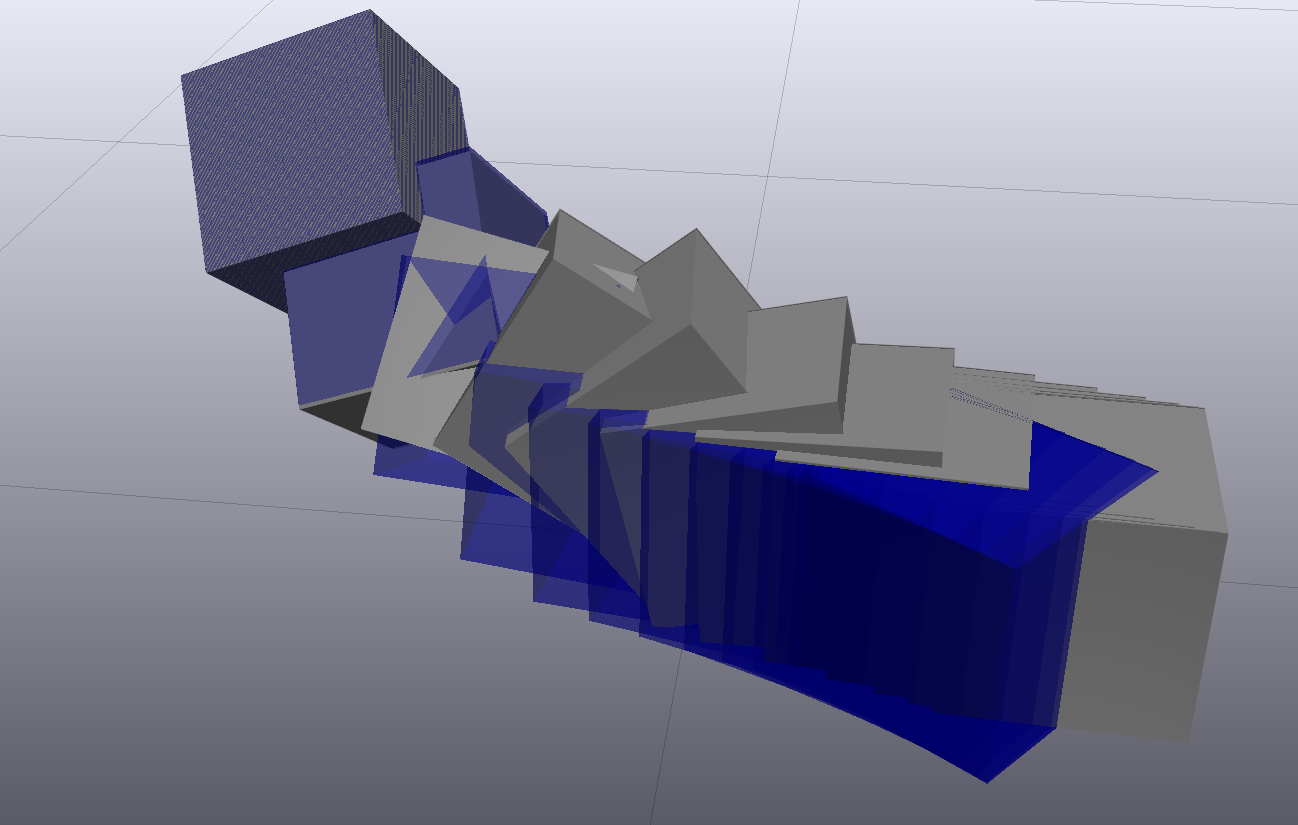}
    \caption{Motion patterns showing the real cube (grey) and the simulated cube (blue) for an inelastic (top) and elastic (bottom) collision in Drake.}
    \label{fig:cube_toss_series}
\end{figure}

\subsection{Cube}
\label{section:cube-discussion}
\subsubsection{Sources of Error} 
\begin{figure}[h]
   	   	\vspace{0.1in}
        \centering
        \includegraphics[width=0.48\textwidth]{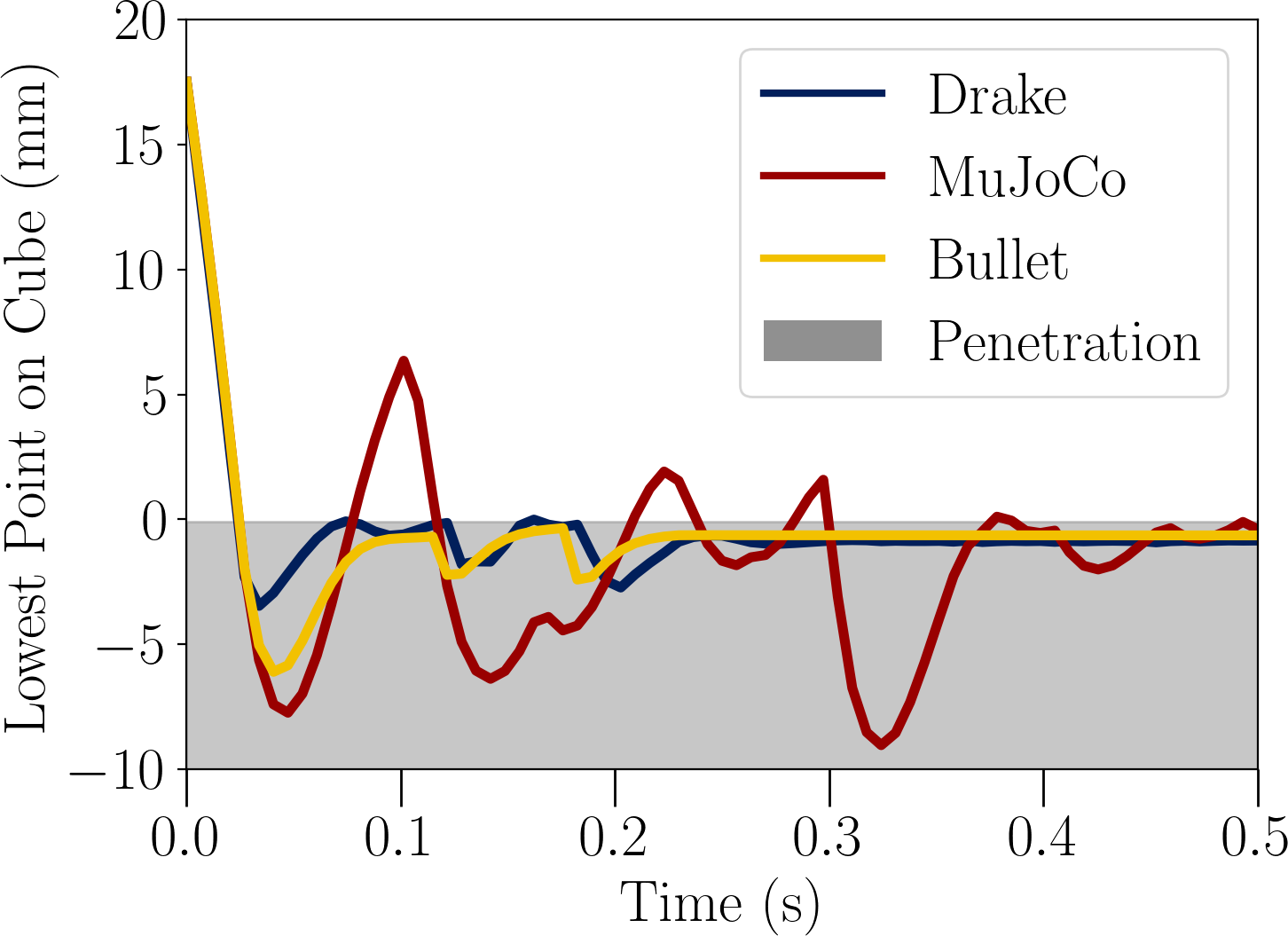}
        \caption{The lowest point on the cube for the three simulators on a typical toss, simulated using the optimal parameters for each simulator. When the lowest point of the cube is below the ground (\textless 0), the system is in penetration. Hand-tuning MuJoCo's damping parameter can reduce the ``bouncing" for this particular toss, but as seen in Figure \ref{fig:cube_sensitivity}, there is no parameter change which could improve the MuJoCo's performance across all logs.}
        \label{fig:energy}
    \end{figure}

 Drake and Bullet have similar errors, as they both recreated near inelastic collision observed in the cube toss system, though they produced less restitution than observed in the real system. For most tosses this resulted in a minor contribution to the average position error, though occasionally caused the simulated cube to slide rather than bounce and settle on a different face (Figure \ref{fig:cube_toss_series}). High rotation error in these instances is responsible for the large standard deviations in Table \ref{tab:cube_errors}. MuJoCo exhibited more elastic behavior, though this was accompanied by softer impacts and larger penetration depths (Figure \ref{fig:energy}). 

\subsubsection{Simulator Comparison}
As seen in Table \ref{tab:cube_errors}, MuJoCo was the least accurate in matching cube trajectories. Due to the interplay between friction and contact dynamics, MuJoCo produces longer contact events with lower peak forces, dissipating energy more slowly than the other simulators. While this effect can be tuned out for individual logs, we found no parameter change which was able to improve the average $e_{q}$ across all logs. 
    
    \subsection{Cassie}
    
\begin{figure}[h]
    	\vspace{0.1in}
        \centering
        \includegraphics[width=0.48\textwidth]{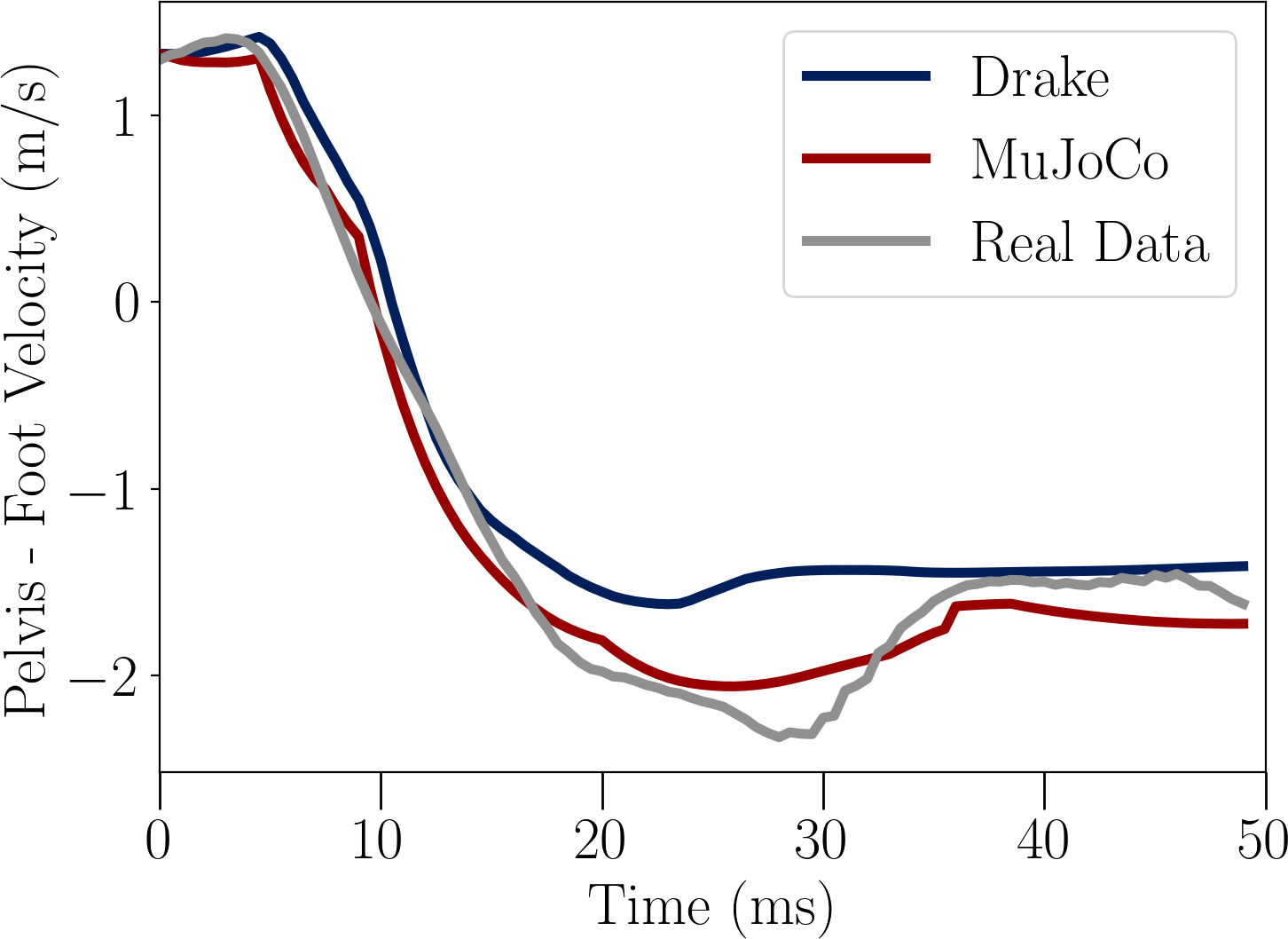}
        \caption{The vertical pelvis velocity of Cassie with respect to its feet for the 50ms duration. In this direction, which is composed from the combined kinematics of the load-bearing joints, both simulators do a remarkable job of capturing the real data. This level of agreement is present in many of the tested logs.}
        \label{fig:cassie_pelvis_velocity}
    \end{figure}

    The velocities of the primary load-bearing joints (hip pitch, knee, and ankle), are captured well by both simulators.
    As a result, the vertical velocity of pelvis, which is the combination of the load-bearing joints, is similarly captured as shown in Fig. \ref{fig:cassie_pelvis_velocity}.
    This is surprising, because the simulators are not only able to predict the velocity at the end of the time window, but also able to model the rate at which the velocity changes as well.
    This is promising evidence in support of the use of simulators to evaluate the performance of controllers \textit{during} the impact event, when the impact has not fully resolved.
    
    Certain logs, where Cassie landed distinctly with the rear parts of the feet first, have significantly high losses across all contact parameters.
    The additional error is attributed to a twisting motion at the pivot points which caused the orientation of the pelvis to shift quickly.
    While this twisting motion is present in those actual trajectories, the actual motion was not as severe.
    This may be due to the contact patches on the physical robot able to exert moments about the contact normal, whereas in both simulators, contact forces are represented as point contacts and unable to produce moments about the contact point.
    
    \subsection{Sensitivity Analysis}
    In addition to identifying the parameters for each simulator and system, we also perform a sensitivity analysis of the parameters by sweeping each parameter value individually, holding the remaining values at $\theta^{*}$.
    The results of this analysis are shown in Fig. \ref{fig:cube_sensitivity}.

    \subsubsection{Stiffness and Damping}
    It is surprising that MuJoCo appears to be insensitive to small stiffness values, given that stiffness is responsible for enforcing non-penetration. Our hypothesis is that, in the small time window around the impact event, the damping forces provide sufficient contact force for our datasets due to the large impact velocity.  
    
    For the cube toss dataset, the sensitivity of Bullet and MuJoCo to insufficient damping, and the correlation of this dependence with the amount of stiffness, suggests that there is an optimal damping ratio for MuJoCo's soft contact dynamics and Bullet's constraint stabilization for this system. Note that this is not the real damping ratio of the contact dynamics, due to trade-offs between friction and contact dynamics in MuJoCo, and the fact that Bullet only uses a spring-damper law to stabilize the simulation of rigid contact.
    
    \subsubsection{Friction}
    The cube toss prediction error is sensitive to the coefficient of friction, which is unsurprising given the large amount of sliding contact in the dataset. Perhaps more surprising is that every simulator is so sensitive to the friction coefficient while not necessarily being close to the experimentally measured value in Table \ref{tab:cube_physics}. Frictional impacts are difficult to accurately model \cite{payr2005oblique} \cite{halm2019modeling}, and the three contact modeling paradigms explored here may need different values to achieve low prediction error and account for the un-modeled effects of each framework.
    
    The Cassie trajectories did not enter the slip-regime for friction, which is necessary to characterize the friction coefficient.
   	This is why the sensitivity to the friction coefficient is relatively flat.
   	
\section{Conclusions}

We observe that for both systems, the simulators tested are able to reproduce the bulk motion observed during impact. 
This suggests that simulators can indeed be an appropriate tool for controller design and verification in impact-rich settings. 
Additionally, we observe that accurate simulator performance can be achieved by a wide range of contact parameters, which suggests that extensive identification of the contact parameters is not necessary.

While the simulators perform well on a significant portion of the cube dataset, there were several cube tosses that all simulators struggle on.
We were unable find any property that nicely distinguishes these tosses from the rest of the dataset, though we noted that these tosses often involve frictional impact with an edge or corner of the cube.
We believe that further investigating and characterizing these particular cases can result in improved simulator performance.

Although the identified cube parameters are fairly stiff, the identified parameters for Cassie are much softer.
This may be due to the rubber pads on Cassie's feet as well as other mechanical compliance associated with a larger system.
This compliance results in impacts resolving over tens of milliseconds, which is important to model when using simulators for validation or in learning situations.

\section*{Acknowledgements}
We thank Mathew Halm for helpful discussions and assistance with the cube-toss dataset and Alejandro Castro for his insight on simulator contact models.


\bibliographystyle{IEEEtran}
\bibliography{impact_references.bib}

\end{document}